\documentclass[conference]{IEEEtran}
\IEEEoverridecommandlockouts
\usepackage{cite}
\usepackage{amsmath,amssymb,amsfonts}
\usepackage{algorithmic}
\usepackage{graphicx}
\usepackage{textcomp}
\usepackage{xcolor}
\def\BibTeX{{\rm B\kern-.05em{\sc i\kern-.025em b}\kern-.08em
    T\kern-.1667em\lower.7ex\hbox{E}\kern-.125emX}}
\begin{document}

\title{Overcoming Scene Context Constraints for Object Detection in wild using Defilters\\

}

\author{\IEEEauthorblockN{1\textsuperscript{st} Vamshi Krishna Kancharla}
\IEEEauthorblockA{\textit{International Institute of Information Technology, Bangalore} \\
\textit{India}\\
}
\and
\IEEEauthorblockN{2\textsuperscript{nd} Neelam sinha}
\IEEEauthorblockA{\textit{Center for Brain Research} \\
\textit{IISC, India}\\
}
}

\maketitle

\begin{abstract}
This paper focuses on improving object detection performance by addressing the issue of image distortions, commonly encountered in uncontrolled acquisition environments. High-level computer vision tasks such as object detection, recognition, and segmentation are particularly sensitive to image distortion. To address this issue, we propose a novel approach employing an image defilter to rectify image distortion prior to object detection. This method enhances object detection accuracy, as models perform optimally when trained on non-distorted images. Our experiments demonstrate that utilizing defiltered images significantly improves mean average precision compared to training object detection models on distorted images. Consequently, our proposed method offers considerable benefits for real-world applications plagued by image distortion. To our knowledge, the contribution lies in employing distortion-removal paradigm for object detection on images captured in natural settings. We achieved an improvement of 0.562 and 0.564 of mean Average precision on validation and test data. 
\end{abstract}


\section{Introduction}
\label{sec:intro}
Deep learning has made significant progress over the last decade, finding remarkable success in a variety of fields, including computer vision and natural language processing, among others. In computer vision, deep neural networks have proven to be capable of powerful feature representation for images, which has led to major advancements in tasks such as object detection.
Object detection is widely considered to be one of the most fundamental and challenging tasks in computer vision. The ability to classify objects in an image by identifying regions of interest within the image is critical to this task, and locating the region of interest is a crucial factor in object detection.
The technique of object detection is the foundation for a wide variety of additional computer vision tasks, such as instance segmentation, picture captioning, and object tracking, to name a few. Object detection is currently being utilized in a broad variety of applications, including autonomous car driving, robotic vision, and video surveillance.
However, object detection models may not perform well when the input images are corrupted by various types of perturbations. Szegedy\cite{b1} demonstrated that even a minute perturbation that is undetectable to the human eye can have a significant impact on the accuracy of an object detection model. Therefore, it is crucial to perform studies to understand the impacts of such perturbations and develop robust object detection models.
In this paper, we do not focus on creating an object detection model for such perturbations, as this would make the model complex. Instead, we are interested in cleaning the images before sending them through the object detection model. Cleaning the images includes removing different types of distortion from the images. Image defilters have been demonstrated to be effective in eliminating distortion from a given image. In the literature, convolutional neural networks (CNNs) have demonstrated state-of-the-art performance for image distortion removal, including denoising, deblurring, super-resolution, dehazing, low-light enhancement, and more.

\section{RELATED WORK}
\label{sec:format}
Beghdadi2022benchmarking \cite{b2} presents a novel investigation into the performance of object detection algorithms in uncontrolled environments, specifically in the presence of image distortions. The authors argue that existing benchmarks for object detection typically use clean images and do not account for the various distortions that can occur in real-world scenarios. 
To facilitate this study, the authors introduce a new dataset named Object Detection Distortion (ODD), built from the MS-COCO dataset, with 10 different types of distortions such as blur, noise, and compression artifacts with varying levels.

In addition, the authors investigate the impact of different distortion types and levels on object detection performance. The study shows that some distortions, such as blur and noise, have a more significant impact on performance than others, such as brightness and contrast changes. The study aimed to investigate the impact of image quality on the development of object detection (OD) methods. The authors found that training OD models with images containing complex scenarios and different types of distortions can significantly improve their performance. The results showed that training with this distorted dataset improved the robustness of OD models. \\

\begin{figure*}[h]
\centering
\includegraphics[trim=0cm 7cm 0cm 0cm,clip,height=8cm,width=18cm]{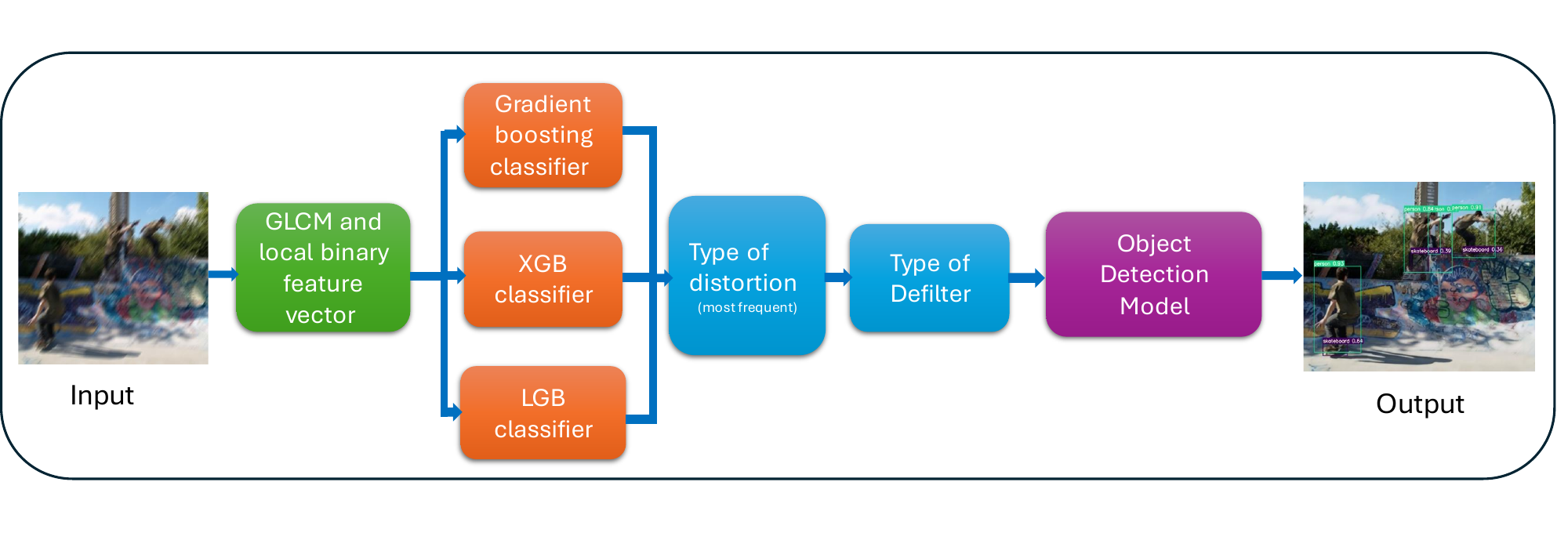}
\caption{Proposed framework pipeline for object detection }
\end{figure*}

The paper "Benchmarking Robustness in Object Detection: Autonomous Driving when Winter is Coming"\cite{b3} proposes a benchmark dataset and evaluation framework for object detection in autonomous driving scenarios, with a focus on winter weather conditions. The authors argue that while existing benchmarks for object detection in autonomous driving are important, they do not adequately capture the challenges of winter driving conditions, which can include factors such as snow, rain, and fog.
To address this gap, the authors introduce a new dataset called the Winter Driving Dataset (WDD), captured in winter conditions across various locations in Canada. The dataset includes annotations for 80 object categories, including cars, pedestrians, and animals.
The authors then use the WDD to evaluate the performance of several state-of-the-art object detection models under winter driving conditions. They find that while some models perform well in clear weather conditions, their performance degrades significantly under winter conditions, highlighting the need for robust object detection models that can handle these challenging scenarios.
Overall, the paper makes an important contribution to the field of object detection in autonomous driving by providing a new dataset and evaluation framework that more accurately captures the challenges of winter driving conditions.

\section{Dataset}
\label{sec:format}
The CD-COCO (Complex Distorted -Common Objects in Context) dataset is a modified version of the well-known MS-COCO dataset, released at ICIP Challenge : Object detection under uncontrolled acquisition environment and scene context constraints, containing 123,000 images divided into three sets: training, validation, and test. In this dataset, specific distortions were applied to the training dataset based on the scene context of each image. The distortion types and severity levels were carefully chosen by considering the object type, position, and scene depth. Local distortions such as blur motion, defocus blur, and backlight illumination applied to objects or specific areas were included in the CD-COCO dataset. Additionally, global distortions caused by camera parameters, acquisition conditions, and atmospheric phenomena like rain and haze were also considered. The authors also took into account the object's position in the observed scene while adjusting the magnitude and weighting of each distortion. This dataset provides a comprehensive evaluation of object detection models' robustness under various types and levels of image distortions in both indoor and outdoor scenes, taking into account the scene specificity and object sensitivity to specific distortions.
Distortion types:(D1) Image Compression, (D2) Noise (Additive White Gaussian Noise), (D3) Contrast changing, (D4) Rain, (D5) Haze, (D6) Motion blur (camera motion), (D7) Defocus Blur, (D8) Local Backlight illumination, (D9) Local Motion Blur, (D10) Local Defocus Blur.
\begin{figure*}[h]
\centering
\includegraphics[trim=0cm 2cm 0cm 0cm,clip,height=13cm,width=20cm]{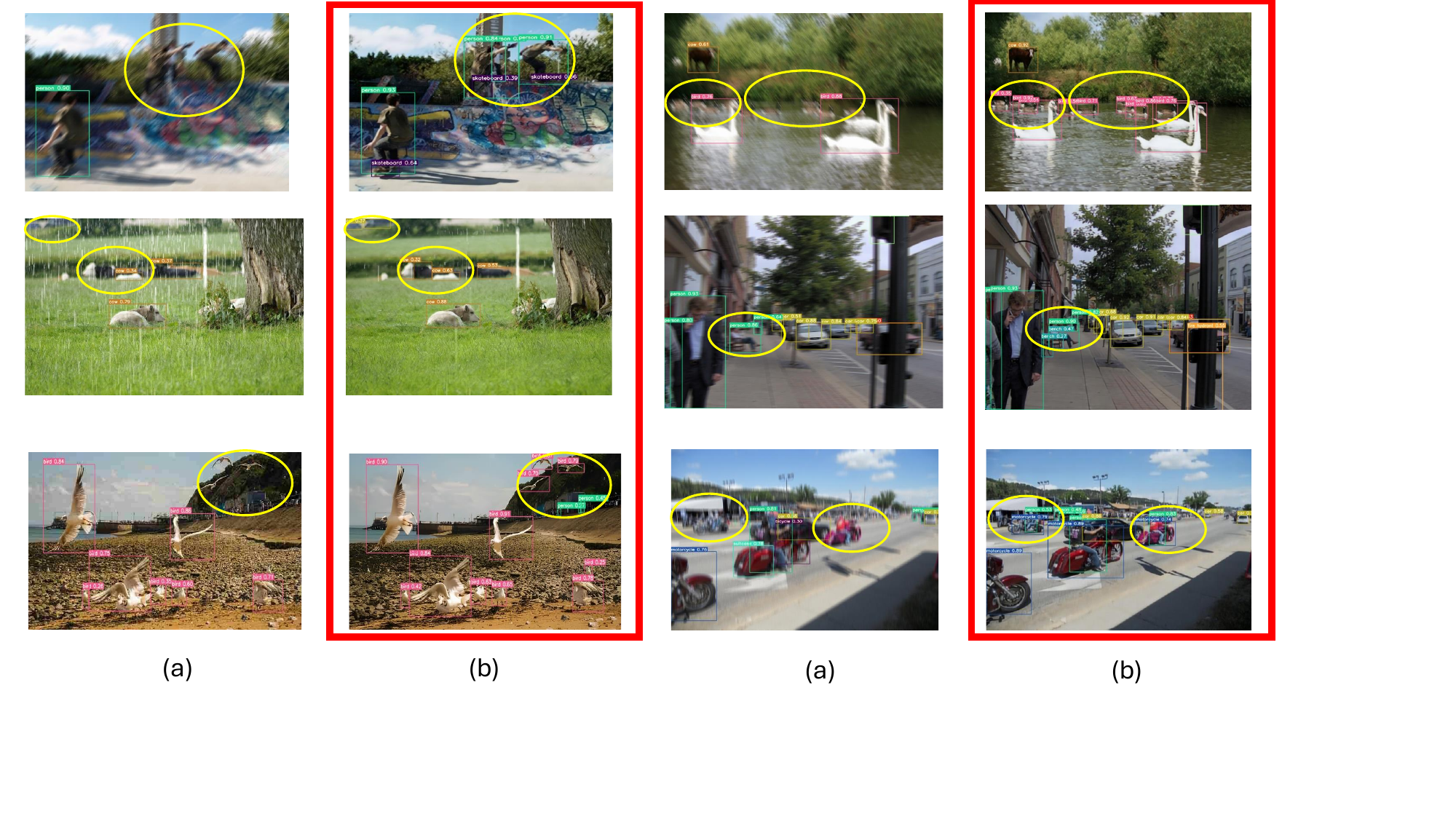}

\caption{(a) Object detection on distorted images, (b) Performance improvement achieved by our proposed method: object detection after defiltering images}
\label{fig:pred}

\end{figure*}
\section{Methodology}
\label{sec:pagestyle}

Firstly, in our proposed methodology, we developed three machine learning classifiers to classify the types of distortion affecting images. The dataset provided for the challenge comprises 10 distinct types of image distortion, which vary in severity and scope from local to global. To facilitate classification, we grouped the different types of distortion into six categories. Specifically, we combined motion and defocus effects with both local and global impacts into a single class, and similarly grouped together images affected by contrast changes and local backlight illumination. The remaining images were classified according to their unique characteristics, resulting in a total of six distinct types of distortion. we calculated the texture properties of a gray level co-occurrence matrix (GLCM), including contrast, dissimilarity, homogeneity, angular second moment, energy, correlation, and local binary pattern, for each image. We then combined the GLCM and local binary patterns into a 58-length feature vector, which we utilized for the classification of the image into six types of distortion using Xgboost, Lightgbm, and Gradient Boosting classifiers. 


\begin{itemize}
    \item \( \mathbf{X} \) as the feature matrix, where each row represents an image and each column represents a feature.
    \item \( y_i \) as the label of the \( i \)th image, indicating the type of distortion.
    \item \( \mathbf{f}_{\text{GLCM}} \) as the feature vector obtained from GLCM properties.
    \item \( \mathbf{f}_{\text{LBP}} \) as the feature vector obtained from Local Binary Patterns.
    \item \( \mathbf{f}_{\text{combined}} \) as the combined feature vector of GLCM and LBP, resulting in a 58-length feature vector.
    \item \( \hat{y}_i \) as the predicted label for the \( i \)th image.
\end{itemize}

Given this, the process can be represented as follows:

\begin{enumerate}
    \item Calculate the GLCM features and Local Binary Pattern features for each image:
    \[
    \mathbf{f}_{\text{GLCM}}^i = \text{GLCM}(i), \quad \mathbf{f}_{\text{LBP}}^i = \text{LBP}(i)
    \]
    \item Combine the GLCM and LBP features into a 58-length feature vector:
    \[
    \mathbf{f}_{\text{combined}}^i = [\mathbf{f}_{\text{GLCM}}^i, \mathbf{f}_{\text{LBP}}^i]
    \]
    \item Train XGBoost, LightGBM, and Gradient Boosting classifiers on the feature matrix \( \mathbf{X} \) and labels \( \mathbf{y} \).
    \item Predict the labels \( \hat{y}_i \) for each image using the trained classifiers.
    \item Take the mode of the predictions across the classifiers as the final predicted label for each image:

\end{enumerate}
\[
\hat{y}_i = \text{mode}(\text{XGBoost}(i), \text{LightGBM}(i), \text{Gradient Boosting}(i))
\]
Secondly, as the challenge dataset did not include any labels describing the different types of distortion, we sourced additional images from open repositories that corresponded to each of these categories. To further refine our classifiers, we also collected one hundred images for each of the six different types of distortion from the training dataset. By leveraging these additional resources, we were able to optimize our machine learning models and achieve robust classification performance across a diverse range of image distortions. This helped us to accurately classify the distortion present in the images and to move onto the next step of the methodology.

Thirdly, after classifying the images, we used image defilters such as derain, deblur, denoise, and dehaze to remove the distortion present in the images. These filters are used to remove unwanted noise, distortions, or artifacts from images, improving their quality and making them easier to analyze or interpret. We employed Maxim\cite{b4} defilters for derain, dehaze, denoise, and deblur, SwinIR\cite{b5} for image compression, and URetinex-Net\cite{b6} for low-light image enhancement. By defiltering the images, we were able to move onto the next step of the methodology, which was object detection.

Finally, we ran the defiltered images through an object detection model to identify the objects present in the image. We tested the defiltered images over two object detection models, yolov7\cite{b7} and InternImage-XL\cite{b8}, and found that Internimage provided the highest mean Average Precision across both the validation and test datasets of 0.562 and 0.564 compared with yolov7 of 0.476 and 0.470. The use of object detection models allowed us to recognize the various objects present in the images and to better understand the information they contained. Figure 2 shows a comparison of object detection over distortion images versus distortion removed images, highlighting the effectiveness of our methodology in improving object detection accuracy.

\section{Results}
\label{sec:pagestyle}
The performance improvement achieved by our proposed strategy is demonstrated in Figure~\ref{fig:pred} and Table 1, which shows the results on the test dataset. In Figure~\ref{fig:pred} Column (a) showcases predictions without applying any post-filtering techniques, representing direct predictions. Column (b) exhibits the predictions generated by the proposed method. In column (a), certain objects that were undetected are highlighted within oval shapes, while in column (b), our proposed methodology successfully detects these objects. Table 1 presents the comparison of our method's performance in terms of mean Average Precision (mAP) on both the validation and test datasets. Our method secured the 2\textsuperscript{nd} position in the challenge, with a mean Average Precision (mAP) of 0.562 over the validation dataset and 0.564 mAP over the test dataset. Notably, our validation and test mAP scores demonstrate consistency, indicating minimal overfitting.
These results highlight the effectiveness of our methodology in improving object detection accuracy and demonstrate its competitive performance in comparison to other approaches.

\begin{table}[]
\begin{tabular}{|l|l|l|}
\hline
\textbf{Team Name}                            & Validation (mAP) & Test (mAP)     \\ \hline
UoM-STAR                              & 0.619            & 0.614          \\ \hline
\textbf{Vamshi Krishna Kancharla}     & \textbf{0.562}   & \textbf{0.564} \\ \hline
AI ZUST Edge Intelligent Security Lab & 0.553            & 0.563          \\ \hline
CUSAT\_AICV                            & 0.574            & 0.550          \\ \hline
EMSE                                  & 0.495            & 0.515          \\ \hline
poderoso54                            & 0.467            & 0.508          \\ \hline
mcleong                               & 0.559            & 0.466          \\ \hline
\end{tabular}
\caption{Comparison of the mean Average Precision (mAP) on the validation and test datasets among different teams}

\end{table}

\section{conclusion}
\label{sec:pagestyle}
In this paper, we introduce a novel approach to address the object detection problem in complex acquisition environments and scene contexts by leveraging image distortion classification and improvement techniques. Our proposed methodology includes the development of machine learning classifiers and the use of image defilters to remove unwanted distortions and artifacts from images. We utilized state-of-the-art de-filters to improve the quality of the images and demonstrated that our approach can improve the mean Average Precision without requiring the model to be trained with distorted data, which would increase its complexity. However, the limited availability of de-filters for local-level distorted images represents a limitation of our study that can be addressed in future research by training a new de-filter with local-level distortion filters. Despite this limitation, our analysis showed promising results and highlights the potential for future studies to improve object detection accuracy through our approach.

The codes will soon be available on GitHub at the following link: https://github.com/kancharlavamshi/icip2023\_challenge

\end{document}